\documentclass[11pt,letterpaper]{article}
\usepackage{naaclhlt2016}
\usepackage{times}
\usepackage{latexsym}
\usepackage{url}
\makeatletter
\newcommand{\@BIBLABEL}{\@emptybiblabel}
\newcommand{\@emptybiblabel}[1]{}
\makeatother
\usepackage[hidelinks]{hyperref}
\usepackage{graphicx}
\usepackage{subfigure}
\usepackage{multirow}
\usepackage{amssymb,amsfonts}
\usepackage{rotating}

\naaclfinalcopy 
\def\naaclpaperid{64} 

\def\figref#1{Figure~\ref{fig:#1}}
\def\figlabel#1{\label{fig:#1}\label{p:#1}}

\def\eqref#1{Eq.~\ref{eqn:#1}}

\def\secref#1{Section~\ref{sec:#1}}

\newcommand{\eversion}{embedding set}
\newcommand{\eversions}{embedding sets}

\newcommand{\EVERSIONS}{Embedding Sets}

\title{Learning Meta-Embeddings \\by Using Ensembles of
  \EVERSIONS}

\author{Wenpeng Yin \rm{and} \textbf{Hinrich Sch\"{u}tze}\\
	    Center for Information and Language Processing\\University of Munich, Germany\\
	    {\tt wenpeng@cis.uni-muenchen.de}}\author{Wenpeng Yin \rm{and} \textbf{Hinrich Sch\"{u}tze}\\
	    Center for Information and Language Processing\\University of Munich, Germany\\
	    {\tt wenpeng@cis.uni-muenchen.de}}

\date{}

\newcounter{notecounter}
\newcommand{\enotesoff}{\long\gdef\enote##1##2{}}
\newcommand{\enoteson}{\long\gdef\enote##1##2{{
\stepcounter{notecounter}
\large\bf
\hspace{1cm}\arabic{notecounter} $<<<$ ##1: ##2
$>>>$\hspace{1cm}}}}
\enoteson
\enotesoff

\begin{document}

\maketitle

\begin{abstract}
Word embeddings -- distributed representations of words --
in deep learning are beneficial for many tasks in natural
language processing (NLP). However, different  \eversions\ vary greatly in quality and characteristics of
the captured semantics. Instead of relying on a more
advanced algorithm for embedding learning, this paper
proposes an ensemble approach of combining different public
\eversions\ with the aim of learning \emph{meta-embeddings}.
Experiments on word similarity and analogy
tasks and on part-of-speech tagging show better performance
of meta-embeddings compared to individual  \eversions.  
One advantage of meta-embeddings is the increased vocabulary coverage.
We will release our
meta-embeddings publicly.
\end{abstract}

\section{Introduction}\label{sec:intro}
Recently, deep neural network (NN) models have achieved remarkable
results in NLP
\cite{collobert2008unified,sutskever2014sequence,rocktaschel2015reasoning}.
\enote{hs}{it's better not to cite yourself -- you can add
  this back in if the paper is accepted}
One reason for these results are \emph{word embeddings},
compact distributed word representations learned in an
unsupervised manner from large corpora
\cite{bengio2003neural,mnih2009scalable,mikolov2010recurrent,mikolov2012statistical,mikolov2013efficient}.

\enote{hs}{cut to save space

Word embeddings are derived by learning a projection of
words from a sparse 1-of-$V$ encoding ($V$: vocabulary
size) onto a lower dimensional and dense vector space and
are treated as feature extractors that encode semantic and
syntactic features of words.

}

Some prior work has studied differences in performance of different
 \eversions.
For example,
\newcite{chen2013expressive} showed that 
the \eversions\
HLBL
\cite{mnih2009scalable}, SENNA \cite{collobert2008unified},
Turian \cite{turian2010word} and Huang
\cite{huang2012improving} have great variance in quality
and characteristics of the semantics captured.
Hill et al. \shortcite{hill2014not,hill2015embedding} showed that embeddings
learned by NN machine translation models can outperform
three representative monolingual  \eversions: word2vec
\cite{mikolov2013distributed}, GloVe
\cite{pennington2014glove} and CW
\cite{collobert2008unified}. \newcite{bansal2014tailoring}
found that Brown clustering, SENNA, CW, Huang and word2vec
yield significant gains 
for dependency parsing. Moreover, using these
representations together achieved the best results, suggesting their complementarity. These prior studies
motivate us to explore an ensemble approach. 
Since each \eversion\ is
trained by a different
NN on a different corpus and can
 be treated  as a distinct
description of words,
our expectation is that the ensemble
contains more information
than each component \eversion. We want to leverage this
\emph{diversity} to learn better-performing word embeddings.

The ensemble approach has two benefits. First,
\emph{enhancement} of the representations: meta-embeddings
perform better than the individual \eversions. Second,
\emph{coverage}: meta-embeddings cover more words than the
individual \eversions. The first three ensemble methods we
introduce are
CONC, SVD and \textsc{1toN} and they
directly only have the benefit of enhancement. They learn meta-embeddings
on the overlapping vocabulary of the \eversions.
CONC concatenates
the  vectors of a word from the different  \eversions. SVD performs
dimension reduction on this concatenation.
\textsc{1toN} assumes that a meta-embedding
for the word exists and uses this meta-embedding to
predict representations of the word in the individual  \eversions\ -- the resulting
fine-tuned meta-embedding is expected to contain knowledge
from all individual  \eversions. 

\enote{hs}{CHECK

I removed
``with
weight balancing.'' because that is a detail that IMHO does
not belong in the intro}

To also address the objective of increased coverage of the vocabulary,
we introduce \textsc{1toN}$^+$, 
a modification of
\textsc{1toN} that 
learns
meta-embeddings for all words in the \emph{vocabulary union} in one
step. 
Let an out-of-vocabulary (OOV) word $w$ of
 \eversion\ ES be a word that is not covered by ES (i.e., ES
does not contain an embedding for $w$).\footnote{We do not
  consider words in this paper that are not covered by any
  of the individual \eversions. OOV always refers to a word
  that is covered by at least one \eversion.}
\textsc{1toN}$^+$ first randomly initializes the
embeddings for  OOVs and the meta-embeddings, then uses a
prediction setup similar to
\textsc{1toN} to update meta-embeddings \emph{as well
  as OOV embeddings}. Thus, \textsc{1toN}$^+$ simultaneously
achieves two goals:
learning meta-embeddings and extending the vocabulary (for
both meta-embeddings and invidual \eversions).

An alternative method that increases coverage is
\textsc{MutualLearning}. \textsc{MutualLearning}
learns the embedding for a word that is an OOV in
 \eversion\  from its embeddings in other  \eversions. 
We will use 
\textsc{MutualLearning} to increase coverage for
CONC, SVD and \textsc{1toN}, so that
these three methods (when used together with 
\textsc{MutualLearning}) have the advantages of both
performance enhancement and increased coverage.

In summary, meta-embeddings 
have two benefits compared to individual \eversions: \emph{enhancement} of performance and
improved \emph{coverage} of the vocabulary.
Below, we demonstrate this experimentally for three tasks: word similarity, word
analogy  and POS tagging.

If we simply view meta-embeddings as a way of coming up with
better embeddings, then the alternative is to develop a
single embedding learning algorithm that produces better
embeddings. Some improvements proposed before
have the disadvantage of increasing the training time of embedding learning
substantially; e.g., the NNLM presented in
\cite{bengio2003neural} is an order of magnitude less
efficient than an algorithm like word2vec and, more
generally, replacing a linear objective function with a
nonlinear objective function increases training time.
Similarly, fine-tuning the hyperparameters of the embedding
learning algorithm is complex and time consuming. In many
cases, it is not possible to retrain using a different
algorithm because the corpus is not publicly available.  But
even if these obstacles could be overcome, it is unlikely
that there ever will be a single ``best'' embedding learning
algorithm. So the current situation of multiple embedding
sets with different properties being available is likely to
persist for the forseeable future.  Meta-embedding learning is a
simple and efficient way of taking advantage of this
diversity.  As we will show below they combine several
complementary embedding sets and the resulting
meta-embeddings are stronger than each individual set.

\section{Related Work}\label{sec:relatedwork}
Related work has focused on improving performance on
specific tasks by using several \eversions\ simultaneously.
To our knowledge, there is no work that aims to learn
generally useful meta-embeddings from individual embedding sets.

\newcite{tsuboi2014neural} incorporated
word2vec and GloVe embeddings into a POS tagging
system and found that using these two  \eversions\ together
was better than using them individually.
Similarly,
\newcite{turian2010word} found that using Brown clusters, CW
embeddings and HLBL embeddings for NER and chunking tasks
together gave better performance than using these
representations individually.

\newcite{luo2014pre} adapted
CBOW \cite{mikolov2013efficient} to train word embeddings on
different datasets -- a Wikipedia corpus,
search click-through data and user query data --
for web search ranking and for word similarity.
They showed that
using these embeddings together gives stronger results than using them individually.

These papers show that using multiple  \eversions\ is
beneficial. However, they either use  \eversions\ trained on
the same corpus \cite{turian2010word} or enhance
 \eversions\ by more training data, not by innovative
learning algorithms \cite{luo2014pre}. In our work, we can
leverage any publicly available  \eversion\ learned by any
learning algorithm. Our meta-embeddings 
are generically useful and are learned by supervised
training of an \emph{explicit model of the dependencies}
between \eversions\ and (except for CONC)
not by \emph{simple concatenation}.

\section{Experimental Embedding Sets}\label{sec:versions}

In this work, we use five released  \eversions. (i) \textbf{HLBL.} Hierarchical log-bilinear 
\cite{mnih2009scalable} embeddings released by
\newcite{turian2010word};\footnote{\url{http://metaoptimize.com/projects/wordreprs/}}
246,122 word embeddings, 100 dimensions; training corpus:
RCV1 corpus (Reuters English newswire, August
1996 -- August 1997). (ii) \textbf{Huang.}\footnote{
  \url{http://ai.stanford.edu/~ehhuang/}}
\newcite{huang2012improving} incorporated global context to
deal with challenges raised by words with multiple
meanings; 100,232 word embeddings, 50 dimensions; training corpus: 
April
2010 snapshot of 
Wikipedia. (iii)
\textbf{GloVe}\footnote{\url{http://nlp.stanford.edu/projects/glove/}} \cite{pennington2014glove}.
1,193,514 word embeddings, 300 dimensions; training corpus: 42 billion tokens of web data, from Common Crawl.
(iv) \textbf{CW} \cite{collobert2008unified}. Released by
\newcite{turian2010word};\footnote{\url{http://metaoptimize.com/projects/wordreprs/}} 
268,810 word embeddings, 200 dimensions; training
corpus:  same as HLBL. (v) \textbf{word2vec}
\cite{mikolov2013distributed}
CBOW;\footnote{\url{http://code.google.com/p/Word2Vec/}} \enote{hs}{do
  we really ned this? The
training was performed using the CBOW architecture, with
sub-sampling using threshold $10^{-5}$, and with negative
sampling with 3 negative examples per each positive one.}  929,022 word
embeddings (we discard phrase embeddings), 300 dimensions; training corpus: Google News (about 100 billion words).

The intersection of the five vocabularies has
size 35,965, the union has size 2,788,636.

\section{Ensemble Methods}\label{sec:ensemble}
This section introduces the four ensemble methods: CONC,
SVD, \textsc{1toN} and \textsc{1toN}$^+$. 


\subsection{CONC: Concatenation}
In CONC, the meta-embedding of $w$  is the
concatenation of five embeddings, one each from the five
\eversions.  For GloVe, we perform L2 normalization
for each dimension across the vocabulary as recommended by
the GloVe authors.  Then each embedding of each
 \eversion\ is L2-normalized. This ensures that each
 \eversion\ contributes equally (a value between -1 and 1) when we compute similarity via dot product.

\begin{figure}[t]
\centering
\includegraphics[width=0.3\textwidth]{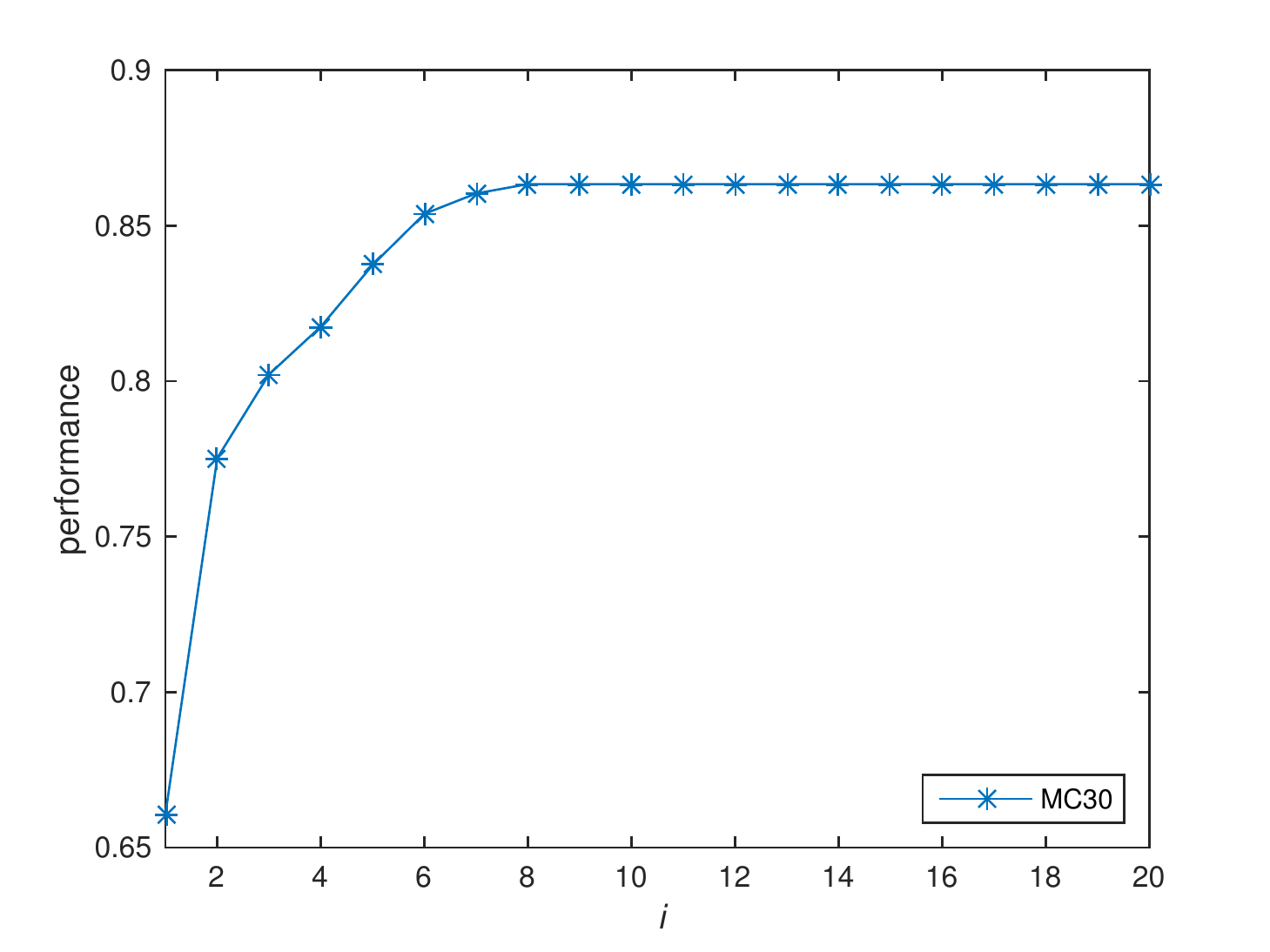}
\caption{Performance vs. Weight scalar $i$}
\label{fig:weight}
\end{figure}

We would like to make use of prior knowledge and
give more weight to well performing
 \eversions.  In this work, we give GloVe and word2vec weight
$i>1$ and weight 1 to the other three  \eversions. 
We use MC30 \cite{miller1991contextual} as dev set, since all
 \eversions\ fully cover it. 
We set $i=8$, the value in  Figure
\ref{fig:weight} where
performance reaches a plateau.
After L2 normalization, GloVe and word2vec
embeddings  are multiplied by $i$ and remaining
 \eversions\ are left unchanged. 


The dimensionality of 
CONC
meta-embeddings is
$k=100+50+300+200+300=950$.

\subsection{SVD: Singular Value Decomposition}
We do SVD on above weighted concatenation vectors of
dimension $k=950$. 

Given a set of CONC representations for $n$
words, each of dimensionality $k$, we compute an SVD
decomposition $C=USV^T$ of the corresponding
$n \times k$ matrix $C$. We then use $U_d$, the first $d$
dimensions of $U$, as the SVD meta-embeddings of the $n$ words.
We apply L2-normalization to embeddings; similarities of SVD
vectors are computed as dot products. 

$d$ denotes the dimensionality of meta-embeddings in
SVD, \textsc{1toN} and \textsc{1toN}$^+$. We use
$d=200$ throughout and investigate the impact of $d$ below.


\subsection{\textsc{1toN}}\label{sec:one2multi}
\begin{figure}[bth]
\centering
\includegraphics[height=2.4cm]{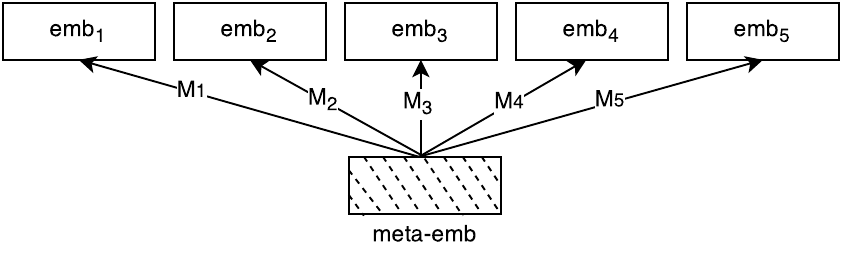}
\caption{1toN}
\label{fig:map2one}
\end{figure}

\figref{map2one} depicts the simple neural network we employ
to learn meta-embeddings in
\textsc{1toN}.  White
rectangles denote known embeddings.
The target to learn is
the meta-embedding (shown as
shaded rectangle).
Meta-embeddings are
initialized randomly.

Let $c$ be the number of  \eversions\  under consideration,
$V_1, V_2, \ldots, V_i, \ldots, V_c$ their vocabularies and
$V^{\cap}=\cap^c_{i=1} V_i$ the intersection,
used as training set. Let $V_*$  denote the meta-embedding space. We define a projection $f_{*i}$ from space
$V_*$ to space $V_i$ ($i=1,2,\ldots,c$) as follows:
\begin{equation}
\mathbf{\hat{w}}_i=\mathbf{M}_{*i}\mathbf{w}_*
\label{equ:proj1}
\end{equation}
where
$\mathbf{M}_{*i}\in\mathbb{R}^{d_i\times d}$,
$\mathbf{w}_*\in\mathbb{R}^d$ is the meta-embedding of
word $w$ in space $V_*$ and 
$\mathbf{\hat{w}}_i\in\mathbb{R}^{d_i}$ is the projected (or learned)
representation of word $w$ in space $V_i$. 
The training objective is minimizing the sum 
of (i) squared error:
\begin{equation}
E=\sum_i |\mathbf{\hat{w}}_i-\mathbf{w}_i|^2
\label{equ:proj2}
\end{equation}
and (ii) L2 cost (sum of squares) of the projection weights
$\mathbf{M}_{*i}$.

As for CONC and
SVD, we weight GloVe and word2vec by $i=8$. For
\textsc{1toN}, we implement this by applying the factor
$i$ to 
the corresponding loss part of the squared error.


The principle of \textsc{1toN} is that we treat each
individual embedding as a projection of the meta-embedding,
similar to
principal component analysis. An embedding is a
description of the word based on the corpus and the model
that were used to create it. The meta-embedding tries to
recover a more comprehensive description of the word when it is
trained to predict
the individual descriptions.

\textsc{1toN} can also be understood as a sentence
modeling process, similar to
DBOW \cite{le2014distributed}. The embedding of each word in
a sentence $s$ is a partial description of $s$.
DBOW combines all partial descriptions to form
a comprehensive description of $s$.  DBOW
initializes the sentence representation randomly, then uses this
representation to predict the representations of 
individual words. 
The sentence representation of $s$ corresponds to the
meta-embedding
in \textsc{1toN}; and 
the representations of the words in $s$
correspond to the
five
embeddings for a word  
in \textsc{1toN}.

\subsection{\textsc{1toN$^+$}}\label{sec:one2multi+}
Recall that an OOV (with respect to \eversion\ ES) is
defined as a word unknown in ES.
\textsc{1toN$^+$} is an extension of
\textsc{1toN} that  learns embeddings for OOVs; thus, it
does
not have the limitation that it can only be run on overlapping vocabulary.

\begin{figure}[tbh]
\centering
\includegraphics[height=2.7cm]{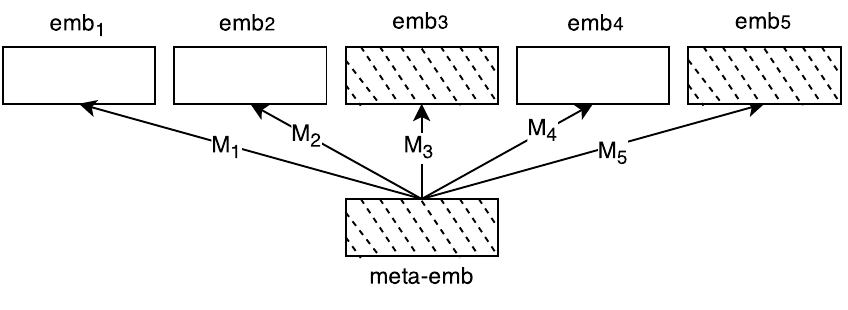}
\caption{1toN$^+$}
\label{fig:map2one_plus}
\end{figure}

Figure \ref{fig:map2one_plus} depicts 
\textsc{1toN$^+$}.  In contrast to Figure
\ref{fig:map2one}, we assume that the current word is an OOV
in  \eversions\ 3 and 5. Hence, in the new
learning task, embeddings 1, 2, 4 are known, and embeddings
3 and 5
and the meta-embedding are targets to learn.

We initialize
all OOV representations and
meta-embeddings randomly and use the
same mapping formula as for \textsc{1toN} to connect
a meta-embedding with the individual embeddings. Both
meta-embedding and initialized OOV embeddings are
updated during training.

\enote{hs}{put back in if there is space

The principle of \textsc{1toN$^+$} is that all information of
known word embeddings will flow to meta-embedding end, and
meta-embeddings can in reverse send knowledge to fine-tune
OOV embeddings. During training, all information will flow
to boost each end until all are ``satisfied''. To be more
direct, the end of meta-embedding acts as a bridge to
connect known embeddings and OOV embeddings, and allows
information to pass through it to learn OOV end, and in
the information flow meta-embedding end gains
knowledge. 
}

Each
 \eversion\ contains  information about only a part of
the overall vocabulary. However, it can predict what the
remaining part should look like by comparing  words
it knows with the information 
other  \eversions\ provide about these words. Thus,
\textsc{1toN$^+$} learns a model of the dependencies
between the individual \eversions\ and can use these
dependencies to infer what the embedding of an OOV should
look like.

CONC, SVD and \textsc{1toN} compute
meta-embeddings only for the intersection
vocabulary.  \textsc{1toN$^+$}
computes meta-embeddings for
the union of all individual vocabularies, thus greatly
increasing the coverage of individual \eversions.

\section{\textsc{MutualLearning}}\label{sec:ml}
\textsc{MutualLearning} is a method
that extends CONC, SVD and \textsc{1toN}
such that they have increased coverage of the vocabulary.
With 
\textsc{MutualLearning},
all four ensemble methods -- CONC, SVD,
\textsc{1toN} and
\textsc{1toN$^+$} -- have the benefits of both performance
enhancement and increased coverage and we can use criteria
like performance, compactness and efficiency of training to select the
best ensemble method for a particular application.

\enote{hs}{reuse if there is space

CONC, SVD and \textsc{1toN} are
supposed to learn high-quality embeddings, but that only
work for overlapping vocabulary (\emph{frequent words:
  appear in all \eversions}) which is only a small part of
the whole vocabulary union. Then,

 how to let the remaining
words that are not covered by five \eversions\ (``\emph{rare
  words}'') share the same strength? This section proposed
an unified framework \textsc{MutualLearning} to learn
representations for rare words, so that i) more words
receive a good representation; ii) the vocabulary of our
final meta-embedding set has a coverage as big as the
vocabulary union of five \eversions.

Adopting the terminology from 
\secref{one2multi}, 
let
$V^{\cup}=\cup^c_{i=1} V_i$ be the vocabulary union.

}

\textsc{MutualLearning} is applied to learn OOV embeddings
for all $c$ \eversions; however,
for ease of exposition, let us assume we want to
compute embeddings for OOVs for \eversion\ $j$ only, based on
known embeddings in the other $c-1$ \eversions, with indexes
$i \in \{ 1 \ldots j-1,j+1 \ldots c\}$. We do this by learning $c-1$
mappings $f_{ij}$, each a projection from 
\eversion\ $E_i$ to 
\eversion\ $E_j$.


Similar to \secref{one2multi}, we train mapping $f_{ij}$ on
the intersection $V_i \cap V_j$ of the vocabularies covered
by the two \eversions. Formally, $\mathbf{\hat{w}}_j=f_{ij}(\mathbf{w}_i)=\mathbf{M}_{ij}\mathbf{w}_i$
where
$\mathbf{M}_{ij}\in\mathbb{R}^{d_j\times d_i}$,
$\mathbf{w}_i\in\mathbb{R}^{d_i}$ denotes the representation of
word $w$ in space $V_i$ and 
$\mathbf{\hat{w}}_j$ is the projected
meta-embedding of word $w$ in space $V_j$. Training loss has
the same form as for \textsc{1toN}. 
A total of  $c-1$ projections $f_{ij}$ are trained to learn
OOV embeddings for \eversion\ $j$.





Let $w$ be a word unknown in the vocabulary $V_j$ of \eversion\ $j$, but  known in
$V_1, V_2, \ldots, V_k$. To compute an embedding for $w$ in
$V_j$, we first compute the $k$ projections
$f_{1j}(\mathbf{w}_1)$, $f_{2j}(\mathbf{w}_2)$, $\ldots$,
$f_{kj}(\mathbf{w}_k)$ from the source spaces $V_1, V_2, \ldots, V_k$ to the target
space $V_j$. Then, the element-wise average of
$f_{1j}(\mathbf{w}_1)$, $f_{2j}(\mathbf{w}_2)$, $\ldots$,
$f_{kj}(\mathbf{w}_k)$ is treated as the representation of $w$ in $V_j$.
Our motivation is that -- assuming there is a
true representation of 
$w$ in $V_j$ and assuming
the projections were learned well -- we would expect 
all the projected vectors
to be close to the true representation. Also,
each
source space contributes potentially complementary
information.  Hence averaging them is a balance of knowledge
from all source spaces.

\section{Experiments}\label{sec:exp}
We train NNs   by
back-propagation 
with AdaGrad \cite{duchi2011adaptive}
and mini-batches.
Table \ref{tab:setup} gives
hyperparameters.

We report results on three tasks:  word similarity, word
analogy and  POS tagging.

\begin{table}[t] 
\small 
\begin{tabular} {l|rlr}
 & \multicolumn{1}{c}{bs} & \multicolumn{1}{c}{lr} & \multicolumn{1}{c}{L2 weight}\\\hline
\textsc{1toN}  & 200 & 0.005 & $5\times 10^{-4}$\\
\textsc{MutualLearning} (ml) & 200 & 0.01 & $5\times 10^{-8}$\\
\textsc{1toN$^+$}  & 2000 & 0.005 & $5\times 10^{-4}$\\
 \end{tabular} 
\caption{Training setup. bs: batch size; lr: learning rate.}\label{tab:setup}
\end{table}

\begin{table*}[htbp]
\setlength{\tabcolsep}{2pt}
  \centering
{\small
  \begin{tabular}{cr|l||rr|rr|rr|rr|rr|rr|rr|r}
    &&Model & \multicolumn{2}{c}{SL999} & \multicolumn{2}{c}{WS353} & \multicolumn{2}{c}{MC30} & \multicolumn{2}{c}{RG}  & \multicolumn{2}{c}{RW} & \multicolumn{2}{|c}{sem.} & \multicolumn{2}{c}{syn.} &tot.\\ \hline \hline
    \multirow{5}{*}{\rotatebox{90}{ind-full}}& 1&HLBL  & 22.1 &(1)& 35.7 &(3) & 41.5 &(0) & 35.2 &(1) &19.1 &(892)& 27.1 &(423) & 22.8 &(198) & 24.7\\
&2    &Huang &  9.7& (3)& 61.7 &(18) & 65.9 &(0) & 63.0 &(0) &6.4 &(982)& 8.4 &(1016) & 11.9 &(326) & 10.4\\
&3    &GloVe & 45.3 &(0)& 75.4 &(18) & 83.6 &(0)& 82.9 &(0)&48.7 &(21)& 81.4 &(0) & 70.1 &(0) & 75.2\\
&4    &CW & 15.6 &(1)& 28.4 &(3) & 21.7 &(0)  & 29.9 &(1) &15.3 &(896)& 17.4 &(423) & 5.0 &(198) & 10.5\\
&5    &W2V & 44.2 &(0)& 69.8 &(0)& 78.9 &(0)& 76.1 &(0)& 53.4 &(209)& 77.1 &(0) & 74.4 &(0) & 75.6\\\hline
    \multirow{5}{*}{\rotatebox{90}{ind-overlap}}& 6&HLBL & 22.3 &(3)&34.8 &(21)&41.5 &(0)&35.2 &(1)&22.2 &(1212)& 13.8 &(8486) & 15.4 &(1859) & 15.4\\
&7    &Huang & 9.7 &(3)&62.0 &(21)&65.9 &(0)&64.1 &(1)&3.9 &(1212)& 27.9 &(8486) & 9.9 &(1859) & 10.7\\
&8   &GloVe &45.0& (3) &75.5 &(21)& 83.6 &(0)& 82.4 &(1)&59.1 &(1212)& 91.1 &(8486) & 68.2 &(1859) & 69.2\\
&9   &CW & 16.0 &(3)&30.8 &(21)&21.7 &(0)&29.9 &(1)&17.4 &(1212)&11.2 &(8486) & 2.3 &(1859)& 2.7\\
&10   &W2V & 44.1 &(3)&69.3 &(21)& 78.9 &(0)& 75.4 &(1)& 61.5 &(1212)& 89.3 &(8486) & 72.6 &(1859) & 73.3\\\hline
    \multirow{20}{*}{\rotatebox{90}{discard}}&11& CONC (-HLBL) & \textbf{46.0}& (3)& \textbf{76.5} &(21)& \textbf{86.3} &(0)& \textbf{82.5} &(1)& \textbf{63.0} &(1211)& \textbf{93.2} &(8486) & \textbf{74.0} &(1859) & \textbf{74.8}\\
&12    & CONC (-Huang) & \textbf{46.1} &(3)& \textbf{76.5} &(21) & \textbf{86.3} &(0)& \textbf{82.5} &(1)& \textbf{62.9} &(1212)& \textbf{93.2} &(8486) & \textbf{74.0} &(1859) & \textbf{74.8}\\
&13    &CONC (-GloVe) & 44.0 &(3)& 69.4 &(21)& 79.1 &(0)& \textbf{75.6} &(1)& 61.5 &(1212)& 89.3 &(8486) & \textbf{72.7} &(1859) & \textbf{73.4}\\
&14    & CONC (-CW) & \textbf{46.0} &(3)& \textbf{76.5} &(21)& \textbf{86.6} &(0)& \textbf{82.5} &(1)& \textbf{62.9} &(1212)& \textbf{93.2} &(8486) & \textbf{73.9} &(1859) & \textbf{74.7}\\
&15    &CONC (-W2V) & 45.0 &(3)& 75.5 &(21)& 83.6 &(0)& 82.4 &(1)& 59.1 &(1212)& 90.9 &(8486) & 68.3 &(1859) & 69.2\\\cline{3-18}
&16&SVD (-HLBL) & \textbf{48.5} &(3)& \textbf{76.1} &(21)& \textbf{85.6} &(0)& \textbf{82.7} &(1)& 61.5 &(1211)& 90.6 &(8486) & 69.5 &(1859) & 70.4\\
&17    &SVD (-Huang) & \textbf{48.8} &(3)& \textbf{76.5} &(21) & \textbf{85.4} &(0)& \textbf{83.5} &(1)& \textbf{61.7} &(1212)& \textbf{91.4} &(8486) & 69.8 &(1859) & 70.7\\ 
&18   &SVD (-GloVe) & \textbf{46.2}& (3)& 66.9 &(21)& 81.6 &(0)& 78.6 &(1)& 59.1 &(1212)& 88.8 &(8486) & 67.3 &(1859) & 68.2\\
&19    &SVD (-CW) & \textbf{48.5} &(3)& \textbf{76.1} &(21)& \textbf{85.7} &(0)& \textbf{82.7} &(1)& 61.5 &(1212)& 90.6 &(8486) & 69.5 &(1859) & 70.4\\
&20   &SVD (-W2V) & \textbf{49.4} &(3)& \textbf{79.0} &(21)& \textbf{87.3} &(0)& 80.7 &(1)& 59.1 &(1212)& 90.3 &(8486) & 66.0 &(1859) & 67.1\\\cline{3-18}
&21&\textsc{1toN} (-HLBL) & \textbf{46.3} &(3)&75.5 &(21) & 82.4 &(0) & 81.0 &(1) & 60.1 &(1211)& \textbf{91.9} &(8486) & \textbf{75.9} &(1859) & \textbf{76.5}\\
&22   &\textsc{1toN} (-Huang) & \textbf{46.5} &(3)&75.4 &(21) & 82.4 &(0) & 82.3 &(1) & 60.2 &(1212)& \textbf{93.5} &(8486) & \textbf{76.3} &(1859) & \textbf{77.0}\\
&23    &\textsc{1toN} (-GloVe) & 43.4 &(3)& 66.5 &(21) & 76.5 &(0) & 75.3 &(1) & 56.5 &(1212)& 89.0 &(8486) & \textbf{73.8} &(1859) & \textbf{74.5}\\
&24   &\textsc{1toN (-CW)} & \textbf{47.4}& (3)&\textbf{76.5} &(21)& \textbf{84.8} &(0) & \textbf{83.4} &(1) & \textbf{62.0} &(1212)& \textbf{91.4} &(8486) & \textbf{73.1} &(1859) & \textbf{73.8}\\
&25    &\textsc{1toN} (-W2V) & \textbf{46.3} &(3)& \textbf{75.9} &(21) & 80.1 &(0) & 78.6 &(1) & 56.8 &(1212)& \textbf{92.2} &(8486) & 72.2 &(1859) & 73.0\\\cline{3-18}

&26&\textsc{1toN$^+$} (-HLBL) & \textbf{46.1}& (3)&\textbf{75.8} &(21) & \textbf{85.5} &(0) & \textbf{83.3} &(1) & \textbf{62.3} &(1211)& \textbf{92.2} &(8486) & \textbf{76.2} &(1859) & \textbf{76.9}\\
&27   &\textsc{1toN$^+$} (-Huang) & \textbf{46.2} &(3)&\textbf{76.1} &(21) & \textbf{86.3} &(0) & \textbf{83.3} &(1)  & \textbf{62.2} &(1212)& \textbf{93.8} &(8486) & \textbf{76.1} &(1859) & \textbf{76.8}\\
&28    &\textsc{1toN}$^+$ (-GloVe)&\textbf{45.3} &(3)& 71.2 &(21) & 80.0 &(0) & 75.7 &(1) & \textbf{62.5} &(1212)& 90.0 &(8486) & \textbf{73.3} &(1859) & \textbf{74.0}\\
&29   &\textsc{1toN$^+$} (-CW) & \textbf{46.9}& (3)&\textbf{78.1} &(21)& \textbf{85.5} &(0) & \textbf{83.9} &(1) & \textbf{62.7} &(1212)& \textbf{91.8} &(8486) & \textbf{73.3} &(1859) & \textbf{74.1}\\
&30    &\textsc{1toN}$^+$ (-W2V) & \textbf{45.8}& (3)& \textbf{76.2} &(21) & \textbf{84.4} &(0) & \textbf{83.1} &(1) & 60.9 &(1212)& \textbf{92.4} &(8486) & 72.4 &(1859) & 73.2\\\hline

    \multirow{4}{*}{\rotatebox{90}{ensemble}}&31 & CONC & \textbf{46.0} &(3)& \textbf{76.5} &(21)& \textbf{86.3} &(0)& \textbf{82.5} &(1)& \textbf{62.9} &(1212)& \textbf{93.2} &(8486) & \textbf{74.0} &(1859) & \textbf{74.8}\\
&32   & SVD & \textbf{48.5}& (3)& \textbf{76.0} &(21)& \textbf{85.7} &(0)& \textbf{82.7} &(1)& 61.5 &(1212)& 90.6 &(8486) & 69.5 &(1859) & 70.4\\
&33    & \textsc{1toN} &\textbf{46.4}& (3)& 74.5 &(21)& 80.7 &(0) & 80.7 &(1) & 60.1 &(1212)& \textbf{91.9} &(8486) & \textbf{76.1} &(1859) & \textbf{76.8}\\
&34    & \textsc{1toN}$^+$ & \textbf{46.3}& (3)& 75.3 &(21)& \textbf{85.2} &(0) & \textbf{82.7} &(1) & \textbf{61.6} &(1212)& \textbf{92.5} &(8486) & \textbf{76.3} &(1859) & \textbf{77.0}
  \end{tabular}
}
\caption{Results on five word similarity tasks and
  analogical reasoning. The number of OOVs is given in
  parentheses for each result.
``ind-full/ind-overlap'': individual
  \eversions\ with respective full/overlapping vocabulary;
  ``ensemble'': ensemble results using all five \eversions;
  ``discard'': one of the five \eversions\ is removed. 
If a result is better than all methods in 
``ind-overlap'', then it is bolded.}\label{tab:wordsimi}
\end{table*}

\begin{table*}[htbp]
\centering
\setlength{\tabcolsep}{2pt}
{\small
\begin{tabular}{cl||rrrr|rrrr|rrrr|rrrr}
&      &       \multicolumn{4}{c|}{RW(21)} & \multicolumn{4}{|c|}{semantic}& \multicolumn{4}{c|}{syntactic}& \multicolumn{4}{c}{total}\\
 &      &   RND & AVG & ml & \textsc{1toN}$^+$ &   RND & AVG & ml & \textsc{1toN}$^+$&   RND & AVG & ml & \textsc{1toN}$^+$&   RND & AVG & ml & \textsc{1toN}$^+$\\\hline\hline
\multirow{3}{*}{\begin{sideways}{ind} \end{sideways}}
    & HLBL &7.4&6.9&17.3&17.5&26.3&26.4&26.3&26.4&22.4&22.4&22.7&22.9&24.1&24.2&24.4&24.5\\
  & Huang &4.4&4.3&6.4&6.4&1.2&2.7&21.8&22.0&7.7&4.1&10.9&11.4&4.8&3.3&15.8&16.2\\
  & CW &7.1&10.6&17.3&17.7&17.2&17.2&16.7&18.4&4.9&5.0&5.0&5.5&10.5&10.5&10.3&11.4\\\hline
\multirow{4}{*}{\begin{sideways}{ensemble} \end{sideways}}
    & CONC &14.2&16.5&48.3&--&4.6&18.0&88.1&--&62.4&15.1&74.9&--&36.2&16.3&81.0&--\\
   & SVD &12.4&15.7&47.9&--&4.1&17.5&87.3&--&54.3&13.6&70.1&--&31.5&15.4&77.9&--\\
   & \textsc{1toN} &16.7&11.7&48.5&--&4.2&17.6&88.2&--&60.0&15.0&76.8&--&34.7&16.1&82.0&--\\
   & \textsc{1toN}$^+$ &--&--&--&48.8&--&--&--&88.4&--&--&--&76.3&--&--&--&81.1
\end{tabular}
}

\caption{Comparison of effectiveness of four methods for learning OOV
  embeddings. 
RND: random
  initialization. 
AVG: average of embeddings of known words.
ml: \textsc{MutualLearning}.
RW(21) means there are still 21 OOVs for the vocabulary union.}\label{tab:ml}
\end{table*}

\subsection{Word Similarity and Analogy Tasks}
We evaluate on  SimLex-999 \cite{hill2015simlex}, WordSim353
\cite{finkelstein2001placing}, RG
\cite{rubenstein1965contextual} and RW
\cite{luong2013better}. 
For completeness, we also show results for
MC30, the validation set.


The word analogy task proposed in
\cite{mikolov2013distributed} consists of questions like,
``$a$ is to $b$ as $c$ is to \_ ?''. The dataset contains
19,544 such questions, divided into a semantic subset of
size 8869 and a syntactic subset of size 10,675.

Table \ref{tab:wordsimi} gives 
\textbf{results on similarity and analogy}. Numbers in
parentheses are line numbers in what follows.
Block ``ind-full'' (1-5) lists the performance of individual
\eversions\ on the \emph{full vocabulary}. Results on lines
6-34 are for the intersection of the vocabularies of the
five embedding sets:
``ind-overlap'' contains the performance of
individual \eversions, ``ensemble''  the performance of our four
ensemble methods and ``discard''
the performance when
one component set is removed.

The four ensemble
approaches are very promising (31-34). For CONC,
discarding HLBL, Huang or CW does not  hurt
performance: CONC (31), CONC(-HLBL) (11),
CONC(-Huang) (12) and CONC(-CW) (14) beat each
individual \eversion\ (6-10) in all tasks.  GloVe contributes most
in SimLex-999, WS353, MC30 and RG;  word2vec contributes most in RW and
word analogy tasks.

SVD (32) reduces the dimensionality of CONC
from 950 to 200, but still gains performance in SimLex-999 and RG.
GloVe
contributes most in SVD (larger losses on line 18 vs.\ lines
16-17, 19-20).
Other embeddings contribute  inconsistently.

\textsc{1toN} performs well only on word analogy, but it
gains great improvement when discarding CW
embeddings (24). \textsc{1toN$^+$} performs better than
\textsc{1toN}: it has stronger results when considering all
\eversions, and can still outperform individual
\eversions\ while discarding HLBL (26), Huang (27) or CW (29).

These results demonstrate that
ensemble methods using multiple \eversions\ produce
stronger embeddings. However, it does not
mean the more \eversions\ the better. Whether an
\eversion\ helps, depends on the complementarity among the sets as well as
how we measure the ensemble results.

CONC, the simplest ensemble, has robust
performance. However, using embeddings of size 950 as input
may mean too many parameters to tune for deep learning.  The
other three methods -- SVD, \textsc{1toN}, \textsc{1toN$^+$}
-- all have the advantage of smaller dimensionality.  SVD
reduces CONC's dimensionality dramatically and still keeps
competitive performance, especially on word similarity.
\textsc{1toN} is competitive on analogy, but weak on
word similarity.  \textsc{1toN$^+$} performs consistently
strongly on word similarity and analogy.

Not all state-of-the-art results are included in Table
\ref{tab:wordsimi}.  One reason is that a fair comparison is
only possible on the shared vocabulary, so methods without
released embeddings cannot be included.  However, GloVe
and word2vec are widely recognized as state-of-the-art
embeddings. In any case, our main contribution is to
present ensemble frameworks which show that a combination of complementary embedding sets
produces better-performing meta-embeddings.

\newcommand\mywidth{0.323}

\begin{figure*}[t]
  \centering 
  \subfigure[Performance vs.\  $d$ of SVD ]{ 
    \label{fig:subfig:a} 
    \includegraphics[width=\mywidth\textwidth]{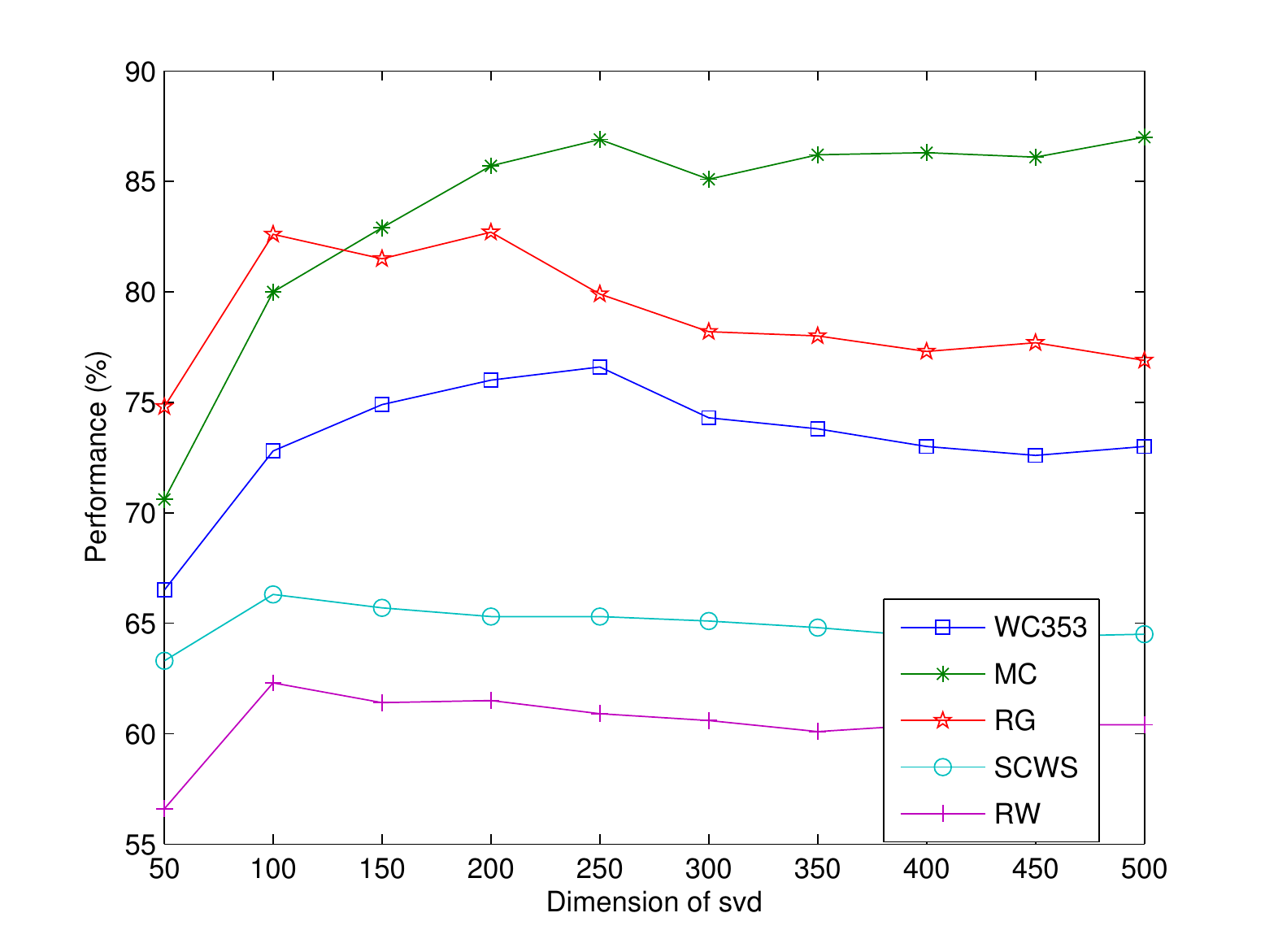}} 
  \subfigure[Performance vs.\  $d$ of \textsc{1toN}]{ 
    \label{fig:subfig:b} 
    \includegraphics[width=\mywidth\textwidth]{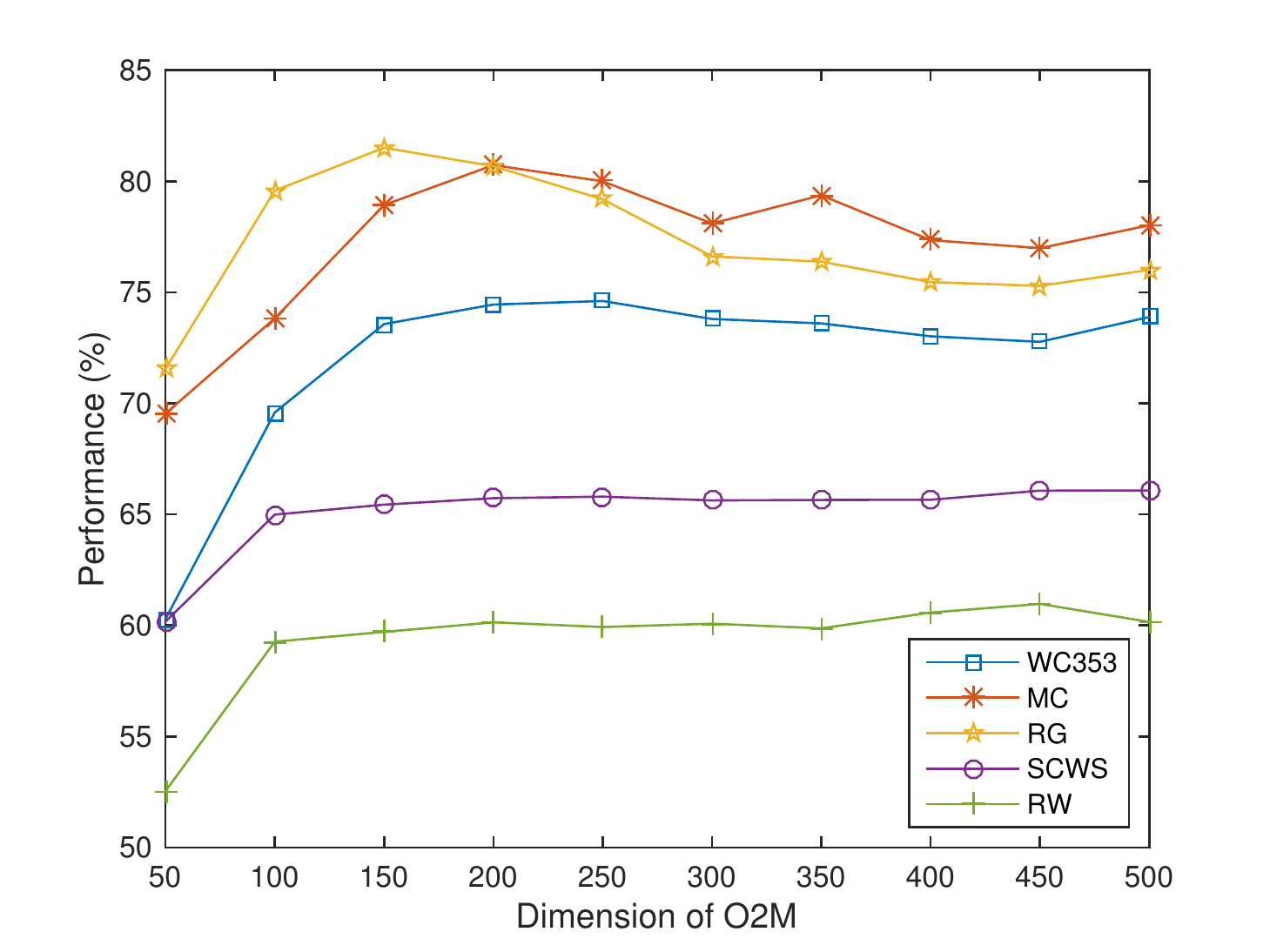}} 
  \subfigure[Performance vs.\  $d$ of \textsc{1toN}$^+$]{ 
    \label{fig:subfig:c} 
    \includegraphics[width=\mywidth\textwidth]{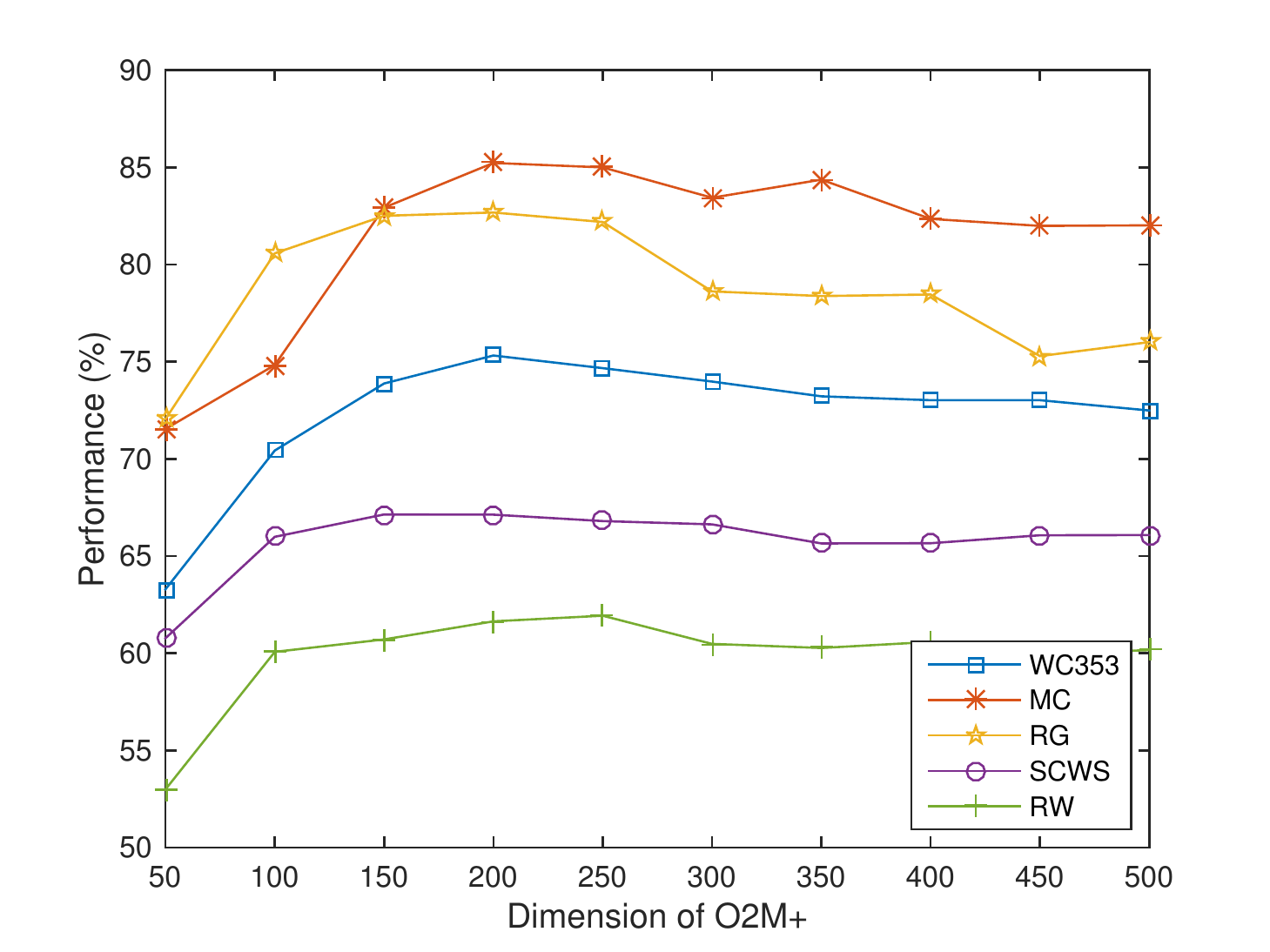}} 
  \caption{Influence of dimensionality} 
  \label{fig:subfig} 
\figlabel{dimd}
\end{figure*}

\begin{table*}[t]
\setlength{\tabcolsep}{3pt}
\centering
{\small
\begin{tabular}{ll|l@{\hspace{0.1cm}}r|l@{\hspace{0.2cm}}r|l@{\hspace{0.2cm}}r|l@{\hspace{0.2cm}}r|l@{\hspace{0.2cm}}r|l@{\hspace{0.1cm}}r}
  &    &   \multicolumn{2}{c|}{newsgroups} & \multicolumn{2}{c|}{reviews}&  \multicolumn{2}{c|}{weblogs} & \multicolumn{2}{c|}{answers}&  \multicolumn{2}{c|}{emails} & \multicolumn{2}{c}{wsj}\\
   &   &   \multicolumn{1}{c}{ALL}&\multicolumn{1}{c|}{OOV}&\multicolumn{1}{c}{ALL}&\multicolumn{1}{c|}{OOV}&\multicolumn{1}{c}{ALL}&\multicolumn{1}{c|}{OOV}&\multicolumn{1}{c}{ALL}&\multicolumn{1}{c|}{OOV}&\multicolumn{1}{c}{ALL}&\multicolumn{1}{c|}{OOV}&\multicolumn{1}{c}{ALL}&\multicolumn{1}{c}{OOV}\\\hline
\multirow{5}{*}{\begin{sideways}{baselines} \end{sideways}}&TnT &88.66&54.73&90.40&56.75&93.33&74.17&88.55&48.32&88.14&58.09&95.76&88.30\\
&Stanford & 89.11 & 56.02 & 91.43 & 58.66 & 94.15 & 77.13 & 88.92&49.30&88.68&58.42&96.83 & 90.25\\
&SVMTool & 89.14 & 53.82 & 91.30 & 54.20 & 94.21 & 76.44 & 88.96&47.25&88.64&56.37&96.63 & 87.96\\
&C\&P & 89.51 & 57.23 & 91.58 & 59.67 & 94.41 & 78.46 & 89.08&48.46&88.74&58.62&96.78 & 88.65\\
&FLORS  &90.86 &66.42&92.95&75.29&94.71&83.64&90.30&62.15&89.44&62.61&96.59&90.37\\\hline

\multirow{5}{*}{\begin{sideways}{+indiv} \end{sideways}}&FLORS+HLBL &90.01&62.64&92.54&74.19&94.19&79.55&90.25&62.06&89.33&62.32&96.53&91.03\\
&FLORS+Huang&90.68&68.53&92.86&77.88&94.71&84.66&90.62&65.04&89.62&64.46&96.65&91.69\\
&FLORS+GloVe&90.99&70.64&92.84&78.19&94.69&86.16&90.54&65.16&89.75&65.61&96.65&92.03\\
&FLORS+CW&90.37&69.31&92.56&77.65&94.62&84.82&90.23&64.97&89.32&65.75&96.58&91.36\\
&FLORS+W2V&90.72&72.74&92.50&77.65&94.75&86.69&90.26&64.91&89.19&63.75&96.40&91.03\\\hline
\multirow{4}{*}{\begin{sideways}{+meta} \end{sideways}}&FLORS+CONC & \textbf{91.87}&72.64&\textbf{92.92}&\textbf{78.34}&\textbf{95.37}&86.69&\textbf{90.69}&\textbf{65.77}&\textbf{89.94}&\textbf{66.90}&\textbf{97.31}&\textbf{92.69}\\
&FLORS+SVD &90.98&70.94&92.47&77.88&94.50&86.49&\textbf{90.75}&64.85&\textbf{89.88}&\textbf{65.99}&96.42&90.36\\
&FLORS+\textsc{1toN} &\textbf{91.53}&\textbf{72.84}&\textbf{93.58}&78.19&\textbf{95.65}&\textbf{87.62}&\textbf{91.36}&\textbf{65.36}&\textbf{90.31}&\textbf{66.48}&\textbf{97.66}&\textbf{92.86}\\
&FLORS+\textsc{1toN}$^+$ &\textbf{91.86}&\textbf{73.36}&\textbf{93.14}&\textbf{78.77}&\textbf{95.65}&\textbf{87.29}&\textbf{91.73}&\textbf{66.28}&\textbf{90.53}&\textbf{66.72}&\textbf{97.75}&\textbf{92.55}\\
\end{tabular}
}
\caption{POS tagging results on six target domains. ``baselines'' lists representative systems for this task, including FLORS. ``+indiv / +meta'': FLORS with individual \eversion\  /  meta-embeddings. Bold means higher than ``baselines'' and ``+indiv''.}\label{tab:pos}
\end{table*}
\textbf{System comparison of learning OOV embeddings.}  In
Table \ref{tab:ml}, we extend the vocabularies of each
individual \eversion\  (``ind'' block) and our ensemble
approaches (``ensemble'' block) to the vocabulary union,
reporting results on RW and  analogy -- these
tasks contain the most
OOVs. As both word2vec and GloVe
have full coverage on  analogy, we do not rereport
them in this table. For each \eversion, we can compute the
representation of an OOV (i) as a randomly
initialized vector (RND); (ii)  as the average of
embeddings of all known words (AVG); (iii) by \textsc{MutualLearning}
(ml) and (iv) by \textsc{1toN}$^+$. 
\textsc{1toN}$^+$ learns 
OOV embeddings for individual
 \eversions\ and meta-embeddings
simultaneously, and it would not make sense to replace these
OOV embeddings computed 
by 
\textsc{1toN}$^+$ with embeddings computed by ``RND/AVG/ml''.
Hence, we do not report 
``RND/AVG/ml'' results for \textsc{1toN}$^+$.

Table \ref{tab:ml} shows four interesting aspects.  (i)
\textsc{MutualLearning} helps much if an \eversion\ has lots
of OOVs in certain task; e.g., \textsc{MutualLearning} is
much better than AVG and RND on RW, and outperforms RND
considerably for CONC, SVD and \textsc{1toN} on analogy.
However, it cannot make big difference for HLBL/CW on
analogy, probably because these two \eversions\ have much
fewer OOVs, in which case AVG and RND work well enough. (ii)
AVG produces bad results for CONC, SVD and \textsc{1toN} on
analogy, especially in the syntactic subtask. We notice that
those systems have large numbers of OOVs in word analogy
task. If for analogy ``$a$ is to $b$ as $c$ is to $d$'', all
four of $a,b,c,d$ are OOVs, then they are represented with
the same average vector. Hence, similarity between $b-a+c$
and each OOV is 1.0. In this case, it is almost impossible
to predict the correct answer $d$.  Unfortunately, methods
CONC, SVD and \textsc{1toN} have many OOVs, resulting in the
low numbers in Table \ref{tab:ml}.  (iii)
\textsc{MutualLearning} learns very effective embeddings for
OOVs. CONC-ml, \textsc{1toN}-ml and SVD-ml all get better
results than word2vec and GloVe on analogy (e.g., for
semantic analogy: 88.1, 87.3, 88.2 vs.\ 81.4 for
GloVe). Considering further their bigger vocabulary, these
ensemble methods are very strong representation learning
algorithms. (iv) The performance of \textsc{1toN}$^+$ for
learning embeddings for OOVs is competitive with
\textsc{MutualLearning}. For HLBL/Huang/CW,
\textsc{1toN}$^+$ performs slightly better than
\textsc{MutualLearning} in all four metrics. Comparing
\textsc{1toN}-ml with \textsc{1toN}$^+$, \textsc{1toN}$^+$
is better than ``ml'' on RW and semantic task, while
performing worse on syntactic task.


\figref{dimd} shows  the
\textbf{influence of
dimensionality}
$d$ for SVD,
\textsc{1toN} and \textsc{1toN$^+$}. 
Peak performance for
different data sets and methods is reached for
$d\in[100,500]$. There are no big differences in the
averages across data sets and methods for high enough $d$,
roughly in the interval $[150,500]$. In summary, as long as
$d$ is chosen to be large enough (e.g., $\ge 150$),
performance is robust.


We will release the meta-embeddings produced by methods 
SVD, \textsc{1toN} and \textsc{1toN$^+$} for
$d=200$ and also the meta-embeddings for method
CONC.

\subsection{Domain Adaptation for POS Tagging}
This section evaluates individual  \eversions\  and  meta-embeddings on POS tagging. FLORS \cite{schnabel2014flors}, the best performing POS tagger for
unsupervised domain adaptation, acts as testbed, in which POS tagging is a window-based, multilabel classification problem using a linear SVM.  A word's  representation consists of \emph{four}
feature vectors based on suffix, shape, left and right distributional neighbors respectively.  We
insert 
word's embedding as the \emph{fifth}
feature vector.
All  \eversions\  (except for
\textsc{1toN}$^+$) are extended to the union vocabulary by
\textsc{MutualLearning}. We follow \newcite{schnabel2014flors} for all feature learning and also train on sections 2-21 of Wall Street Journal (WSJ) and
evaluate on the development sets of six different target
domains: five SANCL \cite{petrov2012overview} domains --
newsgroups, weblogs, reviews, answers, emails -- and
sections 22-23 of WSJ for in-domain testing.

Table \ref{tab:pos} gives results for some representative
systems (``baselines''), FLORS with individual
\eversions\ (``+indiv'') and FLORS with meta-embeddings
(``+meta'').  Following conclusions can be drawn. (i) Not
all individual \eversions\ are beneficial in this task; e.g., HLBL
embeddings make FLORS perform worse in 11 out of 12
cases. (ii) However,  in most cases, embeddings improve 
system performance, which is consistent with prior work on
using embeddings for this type of task
\cite{xiao2013domain,yang2014unsupervised,tsuboi2014neural}. (iii)
Meta-embeddings generally help more than the individual
\eversions, except for SVD (which only performs better
in 3 out of 12 cases).

\section{Conclusion}\label{sec:conc}
This work presented four ensemble methods -- CONC,
SVD, \textsc{1toN} and \textsc{1toN$^+$} --
for learning meta-embeddings from multiple
\eversions. Experiments on word similarity, word analogy
and POS tagging indicated the high quality
of these meta-embeddings. 
The ensemble methods have the added advantage of increasing
vocabulary coverage.
We will release the
meta-embeddings.

\bibliography{naaclhlt2016}
\bibliographystyle{naaclhlt2016}

\end{document}